\documentclass[10pt,twocolumn,letterpaper]{article}

\usepackage{cvpr}
\usepackage{times}
\usepackage{epsfig}
\usepackage{graphicx}
\usepackage{amsmath}
\usepackage{amssymb}

\usepackage{enumitem}
\usepackage{color}
\usepackage{booktabs}
\usepackage{multirow}
\usepackage{subfigure}
\usepackage[table,dvipsnames]{xcolor}

\usepackage[breaklinks=true,bookmarks=false]{hyperref}

\cvprfinalcopy 

\def\cvprPaperID{7560} 

\begin{document}

\title{DEPARA: Deep Attribution Graph for Deep Knowledge Transferability}

\author{Jie Song$^1$\thanks{Equal contribution.}, Yixin Chen$^1$\footnotemark[1], Jingwen Ye$^1$, Xinchao Wang$^2$, Chengchao Shen$^1$, Feng Mao$^3$, 
\\and Mingli Song$^1$
\\$^{1}$Zhejiang University, $^{2}$Stevens Institute of Technology\\
$^{3}$Alibaba Group
}

\maketitle
\thispagestyle{empty}
\pagestyle{empty}  
\begin{abstract}
Exploring the intrinsic interconnections between the knowledge encoded in PRe-trained Deep Neural Networks (PR-DNNs) of heterogeneous tasks sheds light on their mutual transferability, and consequently enables knowledge transfer from one task to another so as to reduce the training effort of the latter. In this paper, we propose the DEeP Attribution gRAph (DEPARA) to investigate the transferability of knowledge learned from PR-DNNs. In DEPARA, nodes correspond to the inputs and are represented by their vectorized attribution maps with regards to the outputs of the PR-DNN. Edges denote the relatedness between inputs and are measured by the similarity of their features extracted from the PR-DNN. The knowledge transferability of two PR-DNNs is measured by the similarity of their corresponding DEPARAs. We apply DEPARA to two important yet under-studied problems in transfer learning: pre-trained model selection and layer selection. Extensive experiments are conducted to demonstrate the effectiveness and superiority of the proposed method in solving both these problems. Code, data and models reproducing the results in this paper are available at \url{https://github.com/zju-vipa/DEPARA}.
\end{abstract}

\section{Introduction}
Driven by massive labeled data~\cite{Deng2009ImageNetAL} and the developing advanced deep models~\cite{He2015DeepRL}, the field of artificial intelligence has made remarkable progress in recent years. However, in real-world scenarios we often encounter the dilemma where limited labeled training data are available for addressing our problems at hand. 
The common practice in this situation is transferring the pre-trained models, which are open sourced by dedicated researchers or industries, to solve our own problems. Yet, along this road comes up another problem: faced with countless PR-DNNs of various layers, which model and which layer of it should be transferred to benefit the target task most? Currently the model selection is usually done blindly by adopting the ImageNet pre-trained models~\cite{Ren2015FasterRT,Long2015FullyCN} and the layer selection is usually conducted heuristically. However, the ImageNet pre-trained models will not always produce satisfactory performances for all the tasks, especially when the task is significantly different from the one defined by ImageNet~\cite{Azizpour2014FactorsOT,Torralba2011UnbiasedLA}. Likewise, the heuristically selected layer may also perform sub-optimally, as the optimal layer for being transferred depends on various factors such as task relatedness and the amount of the target data.

To tackle the aforementioned problems, we need to explore and reveal the underlying transferability among deep knowledge from PR-DNNs of heterogeneous tasks. Recently, Zamir \etal~\cite{Zamir_2018_CVPR} did the pioneering work towards this direction. They proposed a fully computational approach, termed \textit{taskonomy}, to measure the task transferability.
However, there are three unneglected limitations in taskonomy tremendously hampering its real-world application. The first is its prohibitively expensive cost in computation. For computing the pairwise relatedness for a given task dictionary, the computation cost will grow quadratically with the number of the tasks, which will be excessively expensive when the number of tasks becomes large. The second limitation is that it adopts transfer learning to model the relatedness between tasks, which still requires a moderately large amount of labeled data to train the transfer models.
Lastly, taskonomy only consider the transferability across different models or tasks while ignoring the transferability across different layers, which we argue is also important for a transfer to be successful.

The main obstacle standing in the way of measuring the transferability learned from different PR-DNNs is the ``black-box'' nature of deep models. As the knowledge (\textit{e.g.}, features) learned from different PR-DNNs is unexplainable and actually in different embedding space, it is very tricky to compute the transferability directly. In this paper, to derive the transferability of knowledge encoded in PR-DNNs, we propose the DEeP Attribution gRAph (DEPARA) to represent the knowledge learned in PR-DNNs. In DEPARA, nodes correspond to the inputs and are represented by their vectorized attribution maps~\cite{Simonyan2013DeepIC,Bach2015OnPE,Shrikumar2016NotJA} with regards to the outputs of the PR-DNN. Edges denote the relatedness between inputs and are measured by their similarity in the embedding space of the PR-DNN (as seen in Figure~\ref{fig:drg}). As the DEPARAs of different PR-DNNs are defined on the same set of inputs, they are actually in the same embedding space and thus the knowledge transferability of two PR-DNNs is directly measured by the similarity of their corresponding DEPARAs. More similar DEPARAs indicate that more correlated knowledge is learned from different PR-DNNs, thus the knowledge transferability to each other is higher.

The proposed method requires no human annotations, imposes no constraints on architectures and is several-magnitude times faster than taskonomy. Meanwhile, beyond model selection, it can also be easily adopted to the layer selection problem in transfer learning. Extensive experiments conducted demonstrate the effectiveness of DEPARA for quantifying the deep knowledge transferability.

To sum up, we made the following three main contributions: (1) We introduce the challenging, important yet under-studied deep knowledge transferability problem where only PR-DNNs are provided without any labeled data. (2) We propose the DEPARA, an efficient and effective method for deriving the transferability of the knowledge learned from PR-DNNs. To our knowledge, this is the first work to address the pre-trained model selection and the layer selection problems simultaneously. (3) Extensive experiments are conducted to demonstrate the effectiveness of DEPARA in solving both the model and the layer selection problems in transfer learning.

\section{Related Work}
\subsection{Knowledge Transferability}
Transferring PR-DNNs to new tasks is an active research topic. Razavian~\etal \cite{Razavian2014CNNFO} demonstrated that features extracted from deep neural networks could be used as generic image representations to tackle the diverse range of visual tasks. Yosinski~\etal~\cite{Yosinski2014HowTA} investigated the transferability of deep features extracted from every layer of deep neural networks. Azizpour~\etal~\cite{Azizpour2014FactorsOT} studied several factors affecting the transferability of deep features. Recently, the effects of pre-training datasets for transfer learning are also studied~\cite{Kornblith2018DoBI,He2018RethinkingIP,Huh2016WhatMI,Torralba2011UnbiasedLA}. Albeit many heuristics are found by these works, none of them explicitly quantify the transferability among different tasks and layers to provide a principled way for model and layer selection. Zamir~\etal~\cite{Zamir_2018_CVPR} proposed a fully computational approach to measure the task relatedness. Dwivedi and Roig~\cite{Dwivedi2019RepresentationSA} adopted representation similarity analysis for efficient task taxonomy. Song~\etal~\cite{NIPS2019_8849} utilized the similarity of attribution maps to quantify the model transferability. However, the layer selection problem is still omitted in these works. In this paper, we propose DEPARA to address both the model and the layer selection problems in transfer learning.
\subsection{Deep Model Attribution}
Attribution refers to assigning importance scores to the inputs for a specified output. Existing attribution methods can be mainly divided into two groups, including perturbation-~\cite{Zeiler2014VisualizingAU,Zhou2015PredictingEO,Zintgraf2017VisualizingDN} and gradient-based methods~\cite{Simonyan2013DeepIC,Bach2015OnPE,Shrikumar2016NotJA,Sundararajan2017AxiomaticAF,Shrikumar2017LearningIF,Ancona2018TowardsBU}. Perturbation-based methods compute the attribution of an input feature by making perturbations,~\textit{e.g.}, removing, masking or altering, to individual inputs or neurons and observe the impact on later neurons. In contrast, backpropagation-based methods calculate the attributions for all input features in one or few forward and backward pass through the network, which renders them more efficient.
In this paper, we directly adopt existing attribution methods for transferability. Devising more advanced attribution method for our problem is left to future work.
\subsection{Deep Knowledge Representation}
How to represent the knowledge encoded in PR-DNNs is vital for knowledge reusing. Hinton~\etal~\cite{Hinton2015DistillingTK} viewed the soft predictions of a trained teacher model as the knowledge for knowledge distillation.
Following their work, some other forms of knowledge are proposed to facilitate student learning.
For example, Romero~\etal~\cite{Romero2014FitNetsHF} proposed to adopt intermediate representations learned by the teacher as hints to improve the final performance of the student. Zagoruyko and Komodakis~\cite{Zagoruyko2016PayingMA} utilized the attention of the teacher model to guide the learning of the student. Recently, the relation of input instances learned from the trained deep models is also found a kind of useful knowledge~\cite{Chen2017DarkRankAD,Park_2019_CVPR,Liu_2019_CVPR,Tung2019SimilarityPreservingKD,Peng2019CorrelationCF}. For example, Chen~\etal~\cite{Chen2017DarkRankAD}  utilized cross sample similarities to accelerate deep metric learning. Park~\etal~\cite{Park_2019_CVPR} leveraged mutual relations of data examples for knowledge distillation.
In this paper, we propose DEPARA to represent the deep knowledge, which enables us easily quantify the knowledge transferability.

\section{Deep Knowledge Transferability}
\subsection{Notation and Problem Setup}
Assume there are $N$ PR-DNNs available, denoted by $M=\{m_1, m_2, ..., m_N\}$. Each model in $M$ can be viewed to be composed of a number of nonlinear functions: $m_i := f^i_{L_i}\circ\cdots\circ (f^i_2\circ f^i_1)$, where $f$ denotes the basic nonlinear function, $L_i$ denotes the number of nonlinear functions in $m_i$, and the symbol $\circ$ denotes the function composition operation. Note that no constraints are imposed on the architectures of models in $M$, so the number of nonlinear functions in these PR-DNNs may be different. The task handled by $m_i$ is denoted by $t_i$, and all the tasks involved in $M$ are collectively denoted by the task dictionary $T$, $T=\{t_1, t_2, ..., t_N\}$. For task $t_i$, we adopt $P_i(x, y)$ to denote the joint data distribution of the corresponding data domain. In this paper, the term \textit{deep knowledge} refers to the embedding space learned by PR-DNNs. The embedding space produced after $f^i_k$ in $m_i$ is denoted by $\mathcal{F}^i_k$. Given $M$ without any labeled data, we investigate the transferability, which is defined in the next section, between different $\mathcal{F}$s for facilitating task selection and layer selection in transfer learning.
\begin{figure*}[ht]
  \centering
  \includegraphics [scale=0.55]{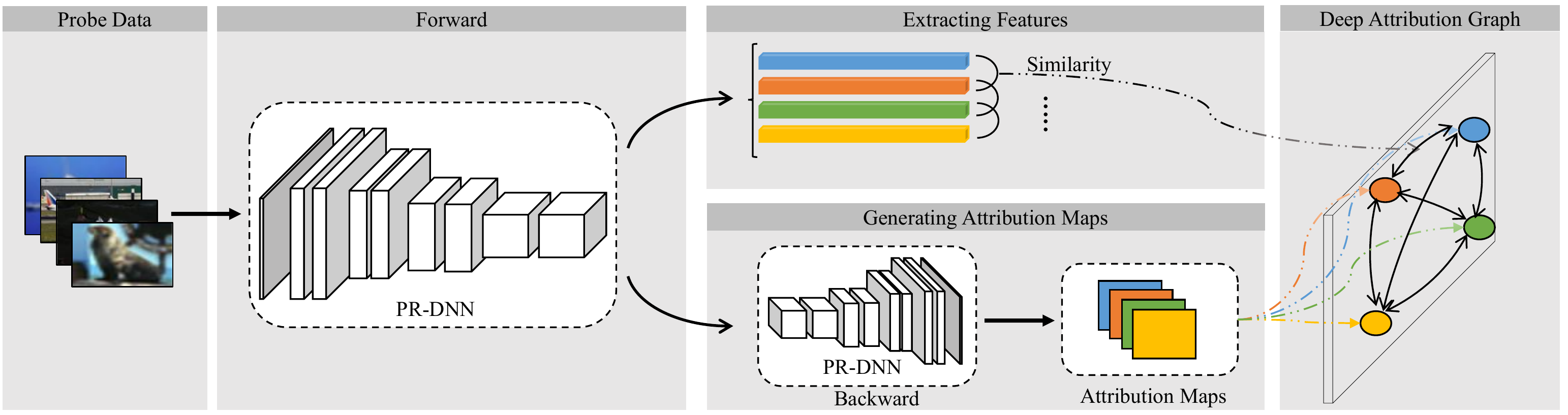}
  \caption{The illustrative diagram of the procedure for constructing the deep attribution graph.}
  \label{fig:drg}
\end{figure*}
\subsection{Definition of Transferability}
An intuitive description of transferability is ``\textit{how well a deep ConvNet representation can be transferred to the target task}''~\cite{Yosinski2014HowTA,Azizpour2014FactorsOT}. Here we introduce a more rigorous definition to facilitate addressing the model and the layer selection problems in transfer learning. Assume there is a deep knowledge pool denoted by $\Omega=\{\mathcal{F}^{(1)},\mathcal{F}^{(2)},...\}$\footnote{Note that we use $\mathcal{F}^{(i)}$ to denote the $i$-th item in $\Omega$, and $\mathcal{F}^{i}$ to denote the knowledge produced by $m_i$.}. Note that in this pool any two knowledge items $\mathcal{F}^{(i)}$ and $\mathcal{F}^{(j)}$ may be produced from different models or layers. The transferability of $\mathcal{F}^{(i)}$ to task $t_j$, denoted by $\mathcal{T}_{\mathcal{F}^{(i)}\rightarrow t_j}$, is defined as the ascending rank of $\mathcal{F}^{(i)}$ among $\Omega$ for solving the target task. Here the rank is computed based on the standard empirical risk. Formally, let $D$ be the target data randomly sampled from $P_j$, \textit{i.e.}, $D = \{(x_1, y_1), (x_2, y_2),...\}$. $\mathcal{F}^{(i)}(D)$ denotes the embeddings of $D$ in $\mathcal{F}^{(i)}$, then
\begin{equation}
\label{eq:transferability}
\mathcal{T}_{\mathcal{F}^{(i)}\rightarrow t_j}(\Omega, D) := ascending\_rank(\mathcal{R}_{P_j}(h_{\mathcal{F}^{(i)}(D)}); \Omega).
\end{equation}
$h_{\mathcal{F}^{(i)}(D)}$ is the hypothesis produced on $\mathcal{F}^{(i)}(D)$. $\mathcal{R}$ denotes the standard expected risk:
\begin{equation}
\mathcal{R}_{P_j}(h) :=  \mathbb{E}_{x,y\sim P_{j}}[\ell_j(h(x), y)],
\end{equation}
where $\ell_j$ is the objective function of task $t_j$. Detailed descriptions of $ascending\_rank$ is provided in the supplementary material. If the transferability of every $\mathcal{F}$ in $\Omega$ to task $t_j$ is known, we can directly select the $\mathcal{F}$ which ranks first for solving the target task $t_j$. Note that when every $\mathcal{F}$ in $\Omega$ comes from a different PR-DNN, the definition of transferability can be used for model selection. If all $\mathcal{F}$s in $\Omega$ come from different layers of the same PR-DNN, the definition can be used for layer selection in transfer learning.

The transferability defined above is intuitively straightforward. However, the computation is expensive for measuring the transferability between every pair of tasks in the task dictionary. What is worse, it needs labeled data for all the tasks involved. To bypass these problems, We propose DEPARA to approximate the defined transferability without any labeled data. We argue two factors must be considered simultaneously for computing the transferability:
\begin{enumerate}
\item  \textbf{Inclusiveness}: for a transfer to be successful, $\mathcal{F}$ produced by the PR-DNN of the source task should be inclusive of sufficient information for solving the target task. Inclusiveness is an intuitively straightforward and fundamental ingredient of transferability. However, since $\mathcal{F}$ is highly nonlinear and unexplainable, it is very challenging to directly measure the inclusiveness of $\mathcal{F}$ for solving the target task.
\item \textbf{Accessibility}: $\mathcal{F}$ should be sufficiently abstracted and easily re-purposed to the target task so that the target task can be well solved with limited human supervision. Without the requirement of accessibility, $\mathcal{F}$ produced by shallower layers will be more likely of higher transferability as $\mathcal{F}$ from shallower layers tend to be of higher inclusiveness than that from higher layers. Measuring the accessibility of $\mathcal{F}$ is also a challenging problem due to the black-box nature of deep models.
\end{enumerate}

\subsection{Deep Attribution Graph}

An illustrative diagram of the DEPARA is depicted in Figure~\ref{fig:drg}.
Formally, assume there is a set of randomly sampled unlabeled data points $D_p=\{x_1, x_2, ..., x_n\}$. $D_p$ is called \textit{probe data} in this paper.
The probe data are first fed to the PR-DNN to obtain their features, \textit{i.e.}, the outputs of the specific layer, after a forward pass. Then the attribution maps are produced by a backward pass. The back-propagation rule depends on the adopted attribution methods~\cite{Ancona2018TowardsBU}. In DEPARA, each node corresponds to a data point in probe data and its feature is the vectorized attribution map of this data point. The edge between two nodes denotes the relatedness of the two data points and are measured by their similarity in the embedding space of the PR-DNN.
For $\mathcal{F}^i_k$ from $m_i$, a DEPARA symbolized by $\mathcal{G}^i_k(D_p)=(\mathcal{V}^i_k, \mathcal{E}^i_k)$ can be obtained, where $\mathcal{V}$ and $\mathcal{E}$ denote the nodes and the edges, respectively. $\mathcal{G}^i_k(D_p)$ indicates the DEPARA is defined on $D_p$. More detailed descriptions of the nodes and the edges are provided as follows.
\subsubsection{Nodes}
The nodes in $\mathcal{G}^i_k$ are collectively denoted by $\mathcal{V}^i_k=\{v^i_{k,1}, v^i_{k,2}, ..., v^i_{k,n}\}$, where $v^i_{k, m}$ is the attribution of $x_m$ with regards to $\mathcal{F}^i_{k}(x_m)$. In this paper, we adopt Gradient*Input~\cite{Shrikumar2016NotJA} for attribution. Gradient*Input refers to a first-order Taylor approximation of how the output would change if the input was set to zero, which implies the importance of the input w.r.t the output. Mathematically, for the $i$-th element $x_{(i)}$ in $x$, its attribution score $v_{(i)}$ with respect to $\mathcal{F}$ is computed as:
\begin{equation}
\label{eq:node}
v_{(i)} :=  x_{(i)}*\frac{\partial{\|\mathcal{F}(x)\|^2}}{\partial{x_{(i)}}},
\end{equation}
where $\|\cdot\|$ denotes $\ell_2$ norm.

The nodes are devised for measuring the inclusiveness of $\mathcal{F}$. The intuition is that for $\mathcal{F}^i(x_m)$ and $\mathcal{F}^j(x_m)$ of the same input $x_m$ but produced by two PR-DNNs $m_i$ and $m_j$, if they produce more similar attributions (\textit{i.e.}, they focus on the more similar regions on the input), they are more likely to contain correlated information and be transformed to each other. Otherwise, they focus on different input dimensions so that being less correlated to each other.

\subsubsection{Edges}
The edges in $\mathcal{G}^i_k$ are collectively denoted by $\mathcal{E}^i_k=\{e^i_{k,11}, e^i_{k,12}, ..., e^i_{k,nn}\}$, where $e^i_{k,pq}$ is the edge of the $p$-th node and the $q$-th node and denotes the similarity of corresponding inputs in the embedding space $\mathcal{F}^i_k$. Formally,
\begin{equation}
\label{eq:edge}
e^i_{k, pq} :=  cosine\_sim(\mathcal{F}^i_{k}(x_p), \mathcal{F}^i_{k}(x_q)).
\end{equation}
We adopt cosine similarity to define the edge because it is insensitive to the length of $\mathcal{F}(\cdot)$. Note that we assume there exists an edge between every pair of nodes in $\mathcal{V}^i_k$, so that $\mathcal{G}^i_k$ is actually a fully connected graph. Furthermore, as $\mathcal{G}^i_k$ is devised to be undirected, $e^i_{k,pq} = e^i_{k,qp}$ for any $p$ and $q$.

The edges are devised to uncover the accessibility of transferability. If the embedding space $\mathcal{F}^i_k$ produced after $f^i_k$ of $m_i$ can be easily transferred (\textit{i.e.}, of high accessibility) to another embedding space $\mathcal{F}^j_l$ produced after $f^j_l$ of $m_j$, $\mathcal{F}^i_k$ and $\mathcal{F}^j_l$ should be similar in topological structure. Otherwise, it will consume a large amount of labeled data and training time to rebuild the embedding space $\mathcal{F}^j_l$ on top of $\mathcal{F}^i_k$, which violates the definition of high accessibility. The edges in $\mathcal{G}$ can be viewed as a representation of the topological structure in the embedding space. Two embedding spaces of the similar topological structure should produce similar edges in $\mathcal{G}$ for the same set of probe data.
\subsection{Task Transferability}
Here we adopt DEPARAs to quantify the transferability among different tasks in $T$, a goal similar to taskonomy~\cite{Zamir_2018_CVPR}. However, in our problem only PR-DNNs of corresponding tasks are provided. We assume no labeled data are available for any task.

Before constructing DEPARAs for the tasks in $T$, two issues must be resolved. The first is that for task $t_i$, which embedding space $\mathcal{F}$ (\textit{i.e.}, layer) of $m_i$ should we choose to best represent the knowledge needed for task $t_i$. In this paper, we viewed all PR-DNNs in an encoder-decoder architecture. The encoder extracts compact features and the decoder makes predictions using the features from the decoder. We adopt the embedding space learned by the encoder, denoted as $\mathcal{F}^i_e$, to represent the knowledge of $t_i$. Thus the knowledge pool can be denoted by $\Omega=\{\mathcal{F}^1_e, \mathcal{F}^2_e, ..., \mathcal{F}^N_e\}$.
The second is that we need a set of probe data which are shared among all the tasks for probing the topological structure of $\mathcal{F}$ and constructing the DEPARAs. In this paper, the probe data are randomly sampled. More details about how the probe data are obtained are provided in the experiment section and the supplementary material.

According to Eq.~(\ref{eq:node}) and~(\ref{eq:edge}), for each task $t$ in $T$, a DEPARA $\mathcal{G}_e$ is obtained on the probe data $D_p$. The transferability of $\mathcal{F}^i_e$ to task $t_j$ is approximated by the descending rank of $\mathcal{F}^i_e$ in $\Omega$ based on the graph similarity:
\begin{equation}
\label{eq:graph_sim}
\mathcal{T}_{\mathcal{F}^i_e\rightarrow t_j}(\Omega, D_p) \approx  descending\_rank(s(\mathcal{G}^i_e, \mathcal{G}^j_e); \Omega, D_p),
\end{equation}
where $s(\cdot)$ is the similarity function. $s(\mathcal{G}^i_e, \mathcal{G}^j_e)=s(\mathcal{V}^i_e, \mathcal{V}^j_e) + \lambda s(\mathcal{E}^i_e, \mathcal{E}^j_e)$. For nodes, we adopt the cosine similarity function: $s(\mathcal{V}^i_e, \mathcal{V}^j_e)=\frac{1}{n}\sum_{k=1}^n{\frac{v^i_{e,k}\cdot v^j_{e,k}}{\|v^i_{e,k}\|\cdot \|v^j_{e,k}\|}}$. For edges, the similarity is defined to be Spearman correlation coefficient: $s(\mathcal{E}^i_e, \mathcal{E}^j_e)= 1 -\frac{6\sum{d_k^2}}{n^3-n}$, where $d_k$ is the difference between the ranks of the $k$-th elements of $\mathcal{E}^i_e$ and $\mathcal{E}^j_e$. $\lambda$ is the trade-off hyper-parameter. Detailed descriptions of $descending\_rank$ are given in the supplementary material.
\subsection{Layer Transferability}
As aforementioned, deep models are usually composed of many nonlinear functions or layers. For a PR-DNN $m_i=f^i_{L_i}\circ\cdots\circ (f^i_2\circ f^i_1)$, actually $L_i$ different embedding spaces can be obtained, which can be denoted by $\Omega_i=\{\mathcal{F}^i_1,\mathcal{F}^i_2,...,\mathcal{F}^i_{L_i}\}$. However, in task transferability described above as well as taskonomy~\cite{Zamir_2018_CVPR}, only one embedding space $\mathcal{F}^i_e$ from the encoder is considered and all other learned knowledge is ignored. It may lead to suboptimal performance as reusing $\mathcal{F}^i_e$ can not guarantee to be optimal for different target tasks.

Here we consider the layer selection problem which is also important in transfer learning: for a source task $t_i$, which layer of its PR-DNN should be choosed to benefit the target task $t_j$ most? The layer selection problem can be viewed as selecting $\mathcal{F}^i$ from $\Omega_i$ which benefits the target task $t_j$ most. We adopt $\mathcal{F}^j_e$ produced by the encoder of $m_j$ to denote the knowledge essential to task $t_j$, as $\mathcal{F}^j_e$ is usually the most compact. The layer selection is conducted by
\begin{equation}
\label{eq:layer_selection}
k =  \arg\max_{k}s(\mathcal{G}^i_k, \mathcal{G}^j_e).
\end{equation}
With $k$ computed from Eq.~(\ref{eq:layer_selection}), we adopt $\mathcal{F}^i_k$ for transferring the PR-DNN $m_i$ to the target task $t_j$.
\section{Experiments}
\begin{figure*}[ht]
  \centering
  \includegraphics [scale=0.45]{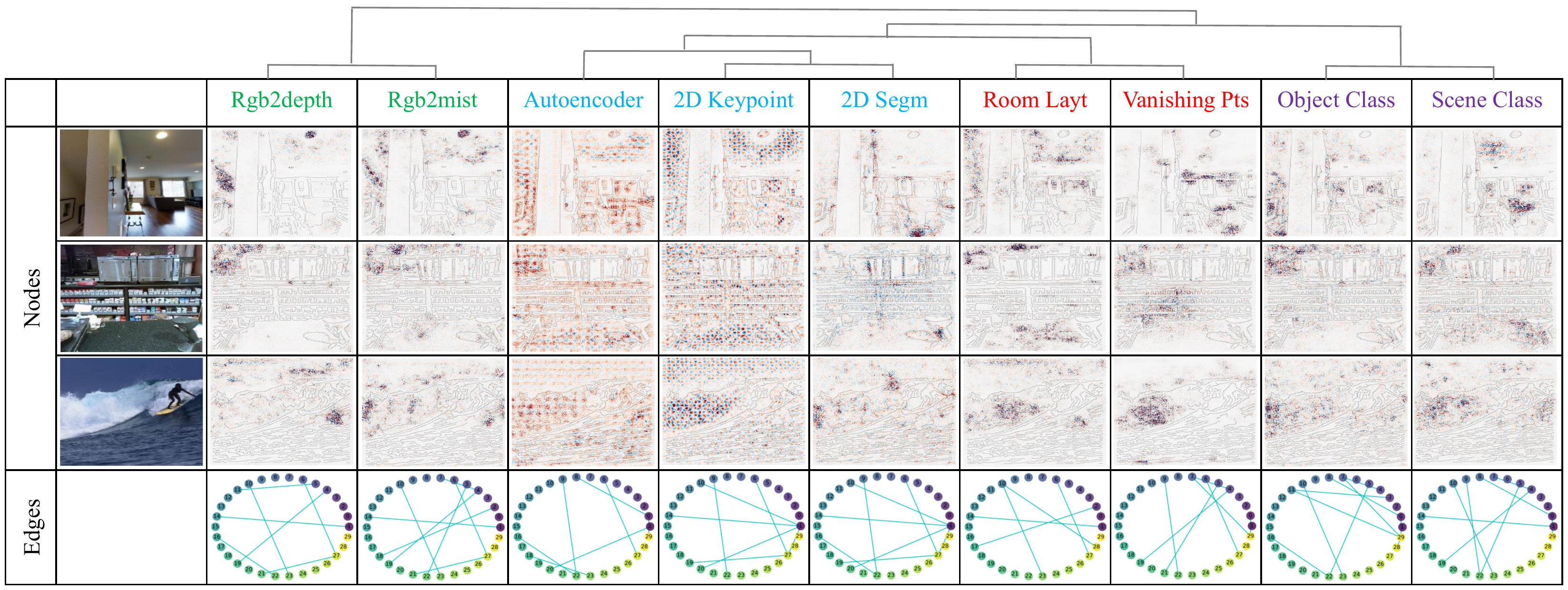}
  \caption{Visualization of some examples of the nodes and the edges of DEPARA. For the nodes, we visualize three examples from taskonomy data, Indoor Scene and COCO, respectively. For the edges, we randomly sample $30$ nodes from taskonomy data and show their interconnections. Note that some weak connections are omitted for better visualization. Here we select two {\color{green}3D} tasks, three {\color{blue}2D} tasks, two {\color{red}geometric} tasks, and two {\color{magenta}semantic} tasks for visualization. The task similarity tree derived from taskonomy is depicted above task names. }
  \label{fig:visualization_tasks}
\end{figure*}
We first validate the proposed method for task transferability, then show its effectiveness for layer selection.
\subsection{Task Transferability on Taskonomy Models}
\subsubsection{Pre-trained Models}
Here we adopt PR-DNNs released by taskonomy~\cite{Zamir_2018_CVPR} to validate the effectiveness of DEPARA for task transferability. Twenty PR-DNNs are selected in this experiment, each of which is for a single-image task. As all taskonomy models naturally follow an encoder-decoder architecture, we directly use the output of the encoder for constructing the DEPARA. Taskonomy measures the task transferability by the performance of transfer learning. We adopt its results to evaluate our method.
\subsubsection{Probe Data}
Following~\cite{NIPS2019_8849}, we construct the probe data by randomly sampling $1,000$ images in the validation set of taskonomy data. We try using more data, but no obvious improvement in performance is observed in our experiment. Additionally, we also adopt Indoor Scene~\cite{Quattoni2009RecognizingIS} and COCO~\cite{Lin2014MicrosoftCC}, which are very different from taskonomy data, as the probe data for computing the transferability of taskonomy tasks. For more details, please refer to the supplementary material.

\subsubsection{Evaluation Metric}
We adopt two evaluation metrics, P@K and R@K, which are widely used in information retrieval, to compare the task transferability constructed from our method with that from taskonomy. Each target task is viewed as a query, and its top-5 source tasks which produce the best transferring performances in taskonomy are regarded as relevant to the query. We adopt the Precision-Recall (PR) curve to demonstrate the overall performance of the proposed method.

\subsubsection{Visualization Results across Tasks}
Here we visualize some nodes in $\mathcal{V}$ and edges in $\mathcal{E}$ of DEPARA to provide a better perceptual understanding of the proposed method. Results are shown in Figure~\ref{fig:visualization_tasks}. It can be seen that some tasks produce similar attribution maps and instance relationships, while some others not. For example, Rgb2depth produces highly similar attribution maps and relational graph with Rgb2mist. However, their results are dissimilar with that of Autoencoder. Actually, Rgb2depth and Rgb2mist are proved in taskonomy of high transferability to each other, while their transferability to Autoencoder is relatively low. Furthermore, taskonomy adopts agglomerative clustering to categorize the tasks into four groups: {\color{green}3D}, {\color{blue}2D}, {\color{red}geometric}, and {\color{magenta}semantic} tasks. From Figure~\ref{fig:visualization_tasks}, we can see that our method tends to produce relatively similar nodes and edges within each group of tasks. Although some exceptions may occur, the results become more credible as we aggregate results of more nodes and edges.
\subsubsection{Task Transferability Results}
\begin{figure}[t]
  \centering
  \subfigure[PR curve.]{
  \begin{minipage}[t]{0.44\linewidth}
  \centering
  \includegraphics[scale=0.12]{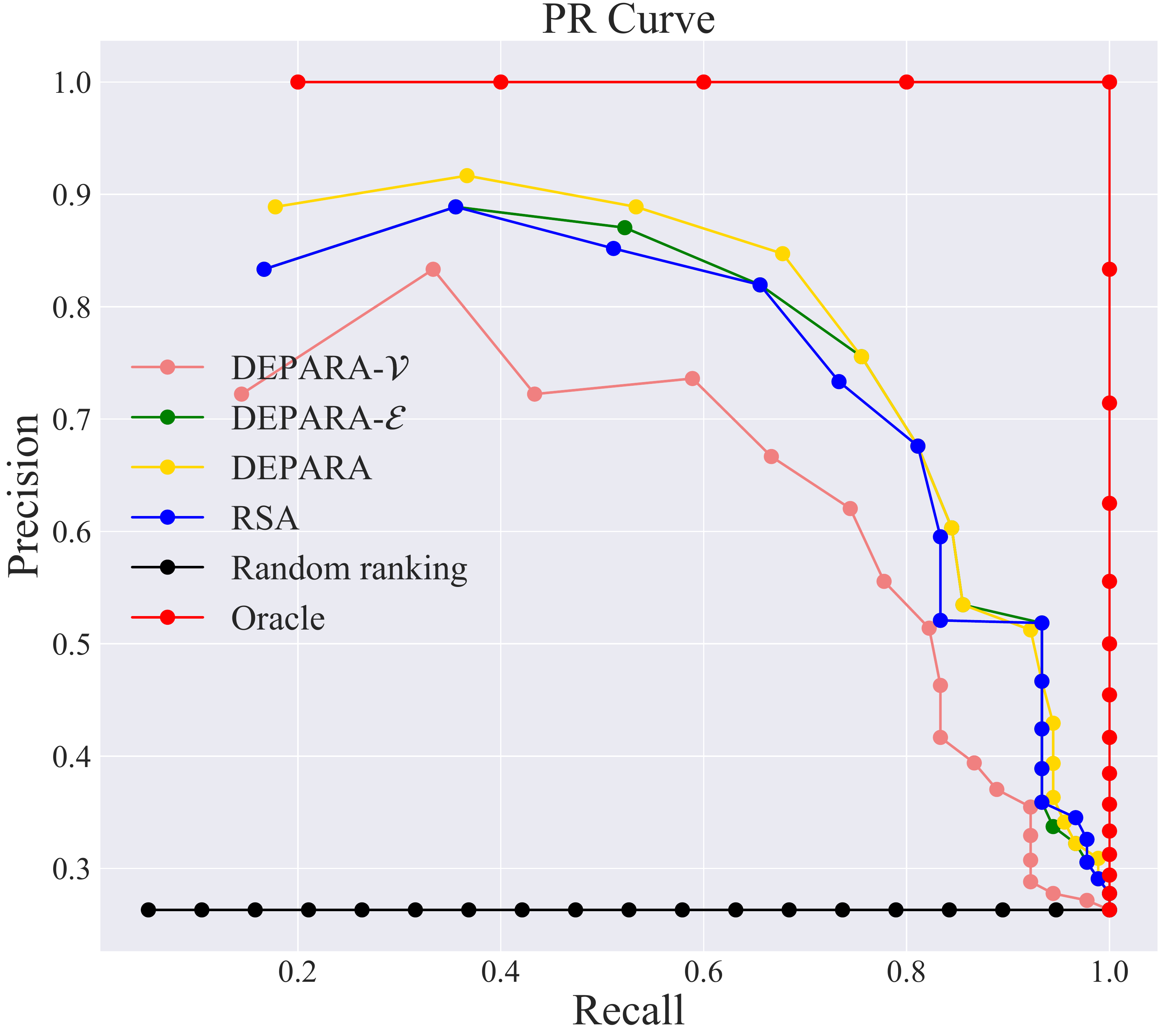}
  \end{minipage}%
  }%
  \subfigure[Task similarity tree.]{
  \begin{minipage}[t]{0.56\linewidth}
  \centering
  \includegraphics[scale=0.24]{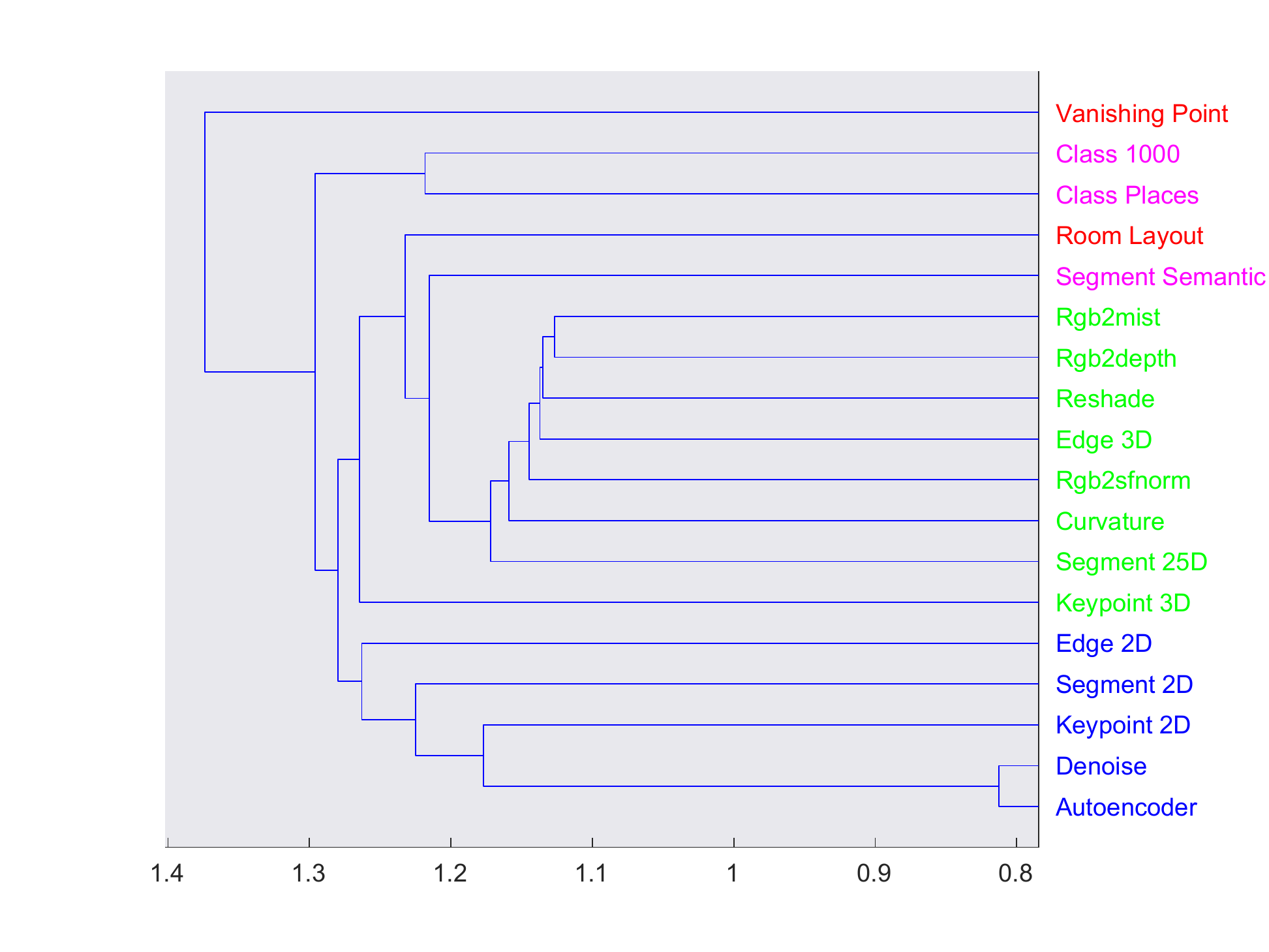}
  \end{minipage}%
  }%
  \caption{PR curve and the task similarity tree obtained on probe data randomly sampled from taskonomy data. }
  \label{fig:precision_recall}
\end{figure}

In this section, we evaluate the proposed method by the task transferability obtained from taskonomy. To better understand the results, we introduce a baseline using Random Ranking, which indicates the task transferability is randomly determined. To make ablation study of the proposed method, we introduce three variants of our method.  DEPARA-${\mathcal{V}}$: only the nodes in DEPARA are utilized for task transferability; DEPARA-${\mathcal{E}}$: only the edges are used; DEPARA: the full version of our method using both nodes and edges, where $\lambda$ is tuned by randomly sampling a small subset of all the PR-DNNs. Additionally, we also introduce another competitor: Representation Similarity Analysis (RSA) proposed by~\cite{Dwivedi_2019_CVPR}. Here we adopt PR curve to compare the performance of all the aforementioned methods. To further demonstrate the similarity between the task transferability obtained by our method and that from taskonomy, the task similarity tree produced by DEPARA is also depicted in Figure~\ref{fig:precision_recall}. The task similarity tree from taskonomy and some other more results are provided in the supplementary material. From these results, we can conclude that:
(1) The proposed method produces task transferability highly similar to that of taskonomy. As our method is much more efficient\footnote{The proposed method takes about $20$ GPU hours on one Quadro P5000 card for pre-trained taskonomy models while taskonomy takes thousands of GPU hours on the cloud for 20 tasks.} than taskonomy, it is an effectual substitute for taskonomy when human annotations are unavailable or the task library $T$ is large in size.
(2) DEPARA outperforms RSA~\cite{Dwivedi_2019_CVPR}, which demonstrates its superiority over the state-of-the-art. Actually, DEPARA-$\mathcal{E}$ and RSA yield comparable performance, as they are quite similar in method.
(3) DEPARA outperforms DEPARA-$\mathcal{V}$ and DEPARA-$\mathcal{E}$ by a considerable margin, which implies the essentiality of both the nodes and the edges in DEPARA for measuring the knowledge transferability. For more results and interesting observations, please refer to the supplementary material.

To investigate the effects of different types of probe data, we also evaluate the proposed method with probe data from Indoor Scene and COCO. The task-wise P@K and R@K results, as well as the average results of the proposed method and some competitors, are provided in Table~\ref{table:task_wise}. It can be seen that although the data from Indoor Scene and COCO are quite different from taskonomy data, the proposed method still produces the task transferability of which the task-wise topological structure is highly similar to the one obtained by taskonomy. It indicates that the proposed method is insensitive to the randomly sampled probe data. Furthermore, the proposed method consistently outperforms DEPARA-$\mathcal{V}$, DEPARA-$\mathcal{E}$ and RSA on all the datasets, which again verifies the effectiveness and superiority of the proposed method.
\begin{table*}[t]
\scriptsize
\begin{center}
\caption{Task-wise similarity between the result from the DEPARA and that from taskonomy. The average results are shown on the right. For a better comparison, average results of DEPARA-$\mathcal{V}$, DEPARA-$\mathcal{E}$ and RSA are also provided.}
\label{table:task_wise}
\begin{tabular}{ccccccccccccccccccc|cccc}
\toprule
&& \rotatebox{90}{AutoEnco} & \rotatebox{90}{Curvature} & \rotatebox{90}{Denoise} & \rotatebox{90}{Edge 2D} & \rotatebox{90}{Edge 3D} & \rotatebox{90}{Keypts 2D} & \rotatebox{90}{Keypts 3D} & \rotatebox{90}{Reshade} & \rotatebox{90}{RGB2Depth} &\rotatebox{90}{RGB2Mist} &\rotatebox{90}{RGB2Norm} &\rotatebox{90}{RoomLayt} &\rotatebox{90}{Segmt 25D} &\rotatebox{90}{Segmt 2D}&\rotatebox{90}{VanishPts} &\rotatebox{90}{SegmtSemt}&\rotatebox{90}{Class1000} &\rotatebox{90}{DEPARA} &\rotatebox{90}{DEPARA-$\mathcal{V}$}&\rotatebox{90}{DEPARA-$\mathcal{E}$} &\rotatebox{90}{RSA}\\
\midrule
\multirow{3}{*}{\rotatebox{90}{Tasknmy}} &P@1 &1.0 &0.0 &1.0 &1.0&0.0&1.0&1.0&1.0&1.0&1.0&1.0&1.0&1.0&1.0&1.0&1.0&1.0&\textbf{0.88}&0.71 &0.82&0.82\\
&P@5 &1.0 &0.6 &1.0 &0.4&0.8&0.8&0.8&0.8&0.8&0.8&0.6&0.8&0.8&0.8&0.8&0.4&0.8&\textbf{0.75}&0.68 &\textbf{0.75}&0.73\\
&R@5 &1.0 &0.6 &1.0 &0.4&0.8&0.8&0.8&0.8&0.8&0.8&0.6&0.8&0.8&0.8&0.8&0.4&0.8&\textbf{0.75}&0.68 &\textbf{0.75}&0.73\\
\midrule
\multirow{3}{*}{\rotatebox{90}{IndoorScn}} &P@1 &1.0 &1.0 &1.0 &1.0&1.0&1.0&1.0&1.0&1.0&1.0&1.0&1.0&1.0&1.0&1.0&1.0&1.0&\textbf{1.00}&0.82 &\textbf{1.00}&\textbf{1.00}\\
&P@5 &1.0 &0.6 &1.0 &0.6&0.6&1.0&1.0&1.0&0.8&0.8&0.8&0.8&0.8&1.0&0.8&0.6&0.6&\textbf{0.81}&0.72 &0.78&0.79\\
&R@5 &1.0 &0.6 &1.0 &0.6&0.6&1.0&1.0&1.0&0.8&0.8&0.8&0.8&0.8&1.0&0.8&0.6&0.6&\textbf{0.81}&0.72 &0.78&0.79\\
\midrule
\multirow{3}{*}{\rotatebox{90}{COCO}} &P@1 &1.0 &0.0 &1.0 &1.0&1.0&1.0&1.0&1.0&1.0&1.0&1.0&1.0&1.0&1.0&0.0&0.0&1.0&\textbf{0.82}&\textbf{0.82} &0.76&\textbf{0.82}\\
&P@5 &1.0 &0.6 &0.8 &0.8&0.6&1.0&1.0&0.8&0.8&1.0&1.0&0.8&1.0&0.8&0.4&0.6&0.6&\textbf{0.80}&0.78 &0.65&0.69\\
&R@5 &1.0 &0.6 &0.8 &0.8&0.6&1.0&1.0&0.8&0.8&1.0&1.0&0.8&1.0&0.8&0.4&0.6&0.6&\textbf{0.80}&0.78 &0.65&0.69\\
\bottomrule
\end{tabular}
\end{center}
\end{table*}

\subsection{Layer Selection in Transfer Learning}
\subsubsection{Experimental Settings} We adopt Syn2Real-C~\cite{Peng2018Syn2RealAN} dataset to validate the effectiveness of DEPARA for layer selection. In Syn2Real-C, the source and the target data are from different domains, but of the same $12$ object categories. The source domain contains $152,397$ synthetic images and the target domain consists of $55,388$ images cropped from the Microsoft COCO dataset. In this paper, we use the data from the source and the target domain to train two domain-specific models. The ultimate goal is improving the performance on the target domain.

We consider two pre-trained models for being transferred to the target: (1) the model trained on the source domain (DNN-Source); (2) the deep model pre-trained on ImageNet (DNN-ImageNet). We adopt the architecture of VGG-19 for both models. DNN-Source is trained from scratch. The initial learning rate is set to be $0.01$ and decayed to $0.001$ after $50$ epochs. We set weight decay to be $0.0005$ and momentum to be $0.9$. DNN-Source is trained for $80$ epochs totally. For DNN-ImageNet, we directly adopt the pre-trained weights provided by TORCHVISION. To compute the transferability of DNN-Source and DNN-ImageNet to classification task on the target domain, we also trained the DNN-Target from scratch on the target data alone.

\subsubsection{Performance of DEPARA for Layer Selection}
Here we show that DEPARA can pick out the layers which yield near the highest performance when transferred to the target task. To this end, we exhaustively conduct the transfer learning for every layer in the pre-trained VGG-19. For each layer transferred to the target task, the current layer and all the layers previous to this layer are fixed and all the layers following the current layer are fine-tuned. As transfer learning usually happens when the target data is scarce, we conduct the experiments in two modes: (1) $0.1$-T: $10\%$ of the target data are used; (2) $0.01$-T: only $1\%$ of the target data are used. In both modes, the pre-trained VGG-19 is further trained for $50$ epochs on target data. To select the transferred layer, we simply set $\lambda$ to be $1$ for both DNN-ImageNet and DNN-Source in $0.1$-T mode. In $0.01$-T mode, as the target data becomes scarcer, the accessibility becomes more important in transferability. Thus we set $\lambda$ to be $10$ in $0.01$-T mode.

Results are listed in Table~\ref{table:layer_wise}. We can see that: (1) The proposed method can successfully pick out the layers which yield the highest performance when transferred to the target. For example, for DNN-ImageNet in $0.01$-T mode, $\#15$, $\#16$, $\#17$ and $\#18$ layers yield the highest transferring performance among all the layers. Our method successfully picks out these layers as they produce the highest DEPARA similarity. Actually, the average Spearman's correlation between the similarity and the accuracy is $0.913$ for all the results shown in Table~\ref{table:layer_wise}, implying that the similarity of DEPARA is a good indicator for layer selection in transfer learning. (2) For different trained models, the layers which yield the highest transferring performance differ. Furthermore, as the size of the target data varies, the best-performing layer may also change. For example, in $0.1$-T mode for DNN-Source, $\#3$, $\#5$ and $\#7$ layers yield the highest performance. However, in $0.01$-T mode the highest-performing layers are  $\#11$, $\#12$ and $\#13$. By appropriately setting $\lambda$, the proposed method can still pick out the best layers for different amounts of target data.
(3) Surprisingly, DNN-ImageNet yields much higher transferring performance than DNN-Source. The similarity of $\mathcal{E}$ in some layers of DNN-ImageNet is significantly higher than that of DNN-Source, which implies that the embedding space learned on ImageNet is more suitable for solving the target task. The DNN-Source, albeit trained on the same task as the target, learned quite a different embedding space due to the large difference between the source and the target domain. Thus it produces relatively worse performance when transferred to the target data.
(4) Trained from scratch on the target data, VGG-19 achieves $61.74\%$ and $32.27\%$ accuracy in $0.1$-T and $0.01$-T mode, respectively. Comparing these figures to the results in Table~\ref{table:layer_wise}, we can see that some layers produce worse performance when transferred to the target data than they are trained from scratch. This phenomenon is known as \textit{negative transfer}~\cite{Wang_2019_CVPR}. Negative transfer occurs especially when the PR-DNN is trained on a quite different domain (like DNN-Source) or for an unrelated task to the target task. For DNN-Source, most layers produce negative transfer when transferred to the target data. All these results imply the importance of both the model selection and the layer selection in transfer learning.
\begin{table*}[ht]
\scriptsize
\begin{center}
\caption{Layer-wise transferring performance of DNN-ImageNet and DNN-Source transferred to the target domain. SIM denotes the similarity between the DEPARAs of the specific layer and the target task. ACC denotes the accuracy on target test data. For space consideration, we omit the $2$-nd, $4$-th, $6$-th and $8$-th layers of VGG-19. Darker color indicates higher values.}
\label{table:layer_wise}
\begin{tabular}{ccccccccccccccccc}
\toprule
&&&\multicolumn{12}{c}{CONVOLUTIONAL LAYERS} &\multicolumn{2}{c}{FC LAYERS}\\
\cmidrule(lr){4-15} \cmidrule(lr){16-17}
&&&$\#1$&$\#3$&$\#5$&$\#7$&$\#9$&$\#10$&$\#11$&$\#12$&$\#13$&$\#14$&$\#15$&$\#16$ &$\#17$&$\#18$\\
\midrule
\multirow{6}{*}{\rotatebox{90}{DNN-ImageNet}} &\multirow{4}{*}{SIM} &$\mathcal{V}$ &\cellcolor{cyan!10}0.45 &\cellcolor{cyan!10}0.45 &\cellcolor{cyan!25}0.48 &\cellcolor{cyan!45}0.52&\cellcolor{cyan!60}0.55&\cellcolor{cyan!60}0.55&\cellcolor{cyan!60}0.55&\cellcolor{cyan!60}0.55&\cellcolor{cyan!55}0.54&\cellcolor{cyan!55}0.54&\cellcolor{cyan!55}0.54&\cellcolor{cyan!55}0.54&\cellcolor{cyan!50}0.53&\cellcolor{cyan!45}0.52\\
&&$\mathcal{E}$ &\cellcolor{blue!16}0.16 &\cellcolor{blue!01}0.01 &\cellcolor{blue!20}0.20 &\cellcolor{blue!03}0.03&\cellcolor{blue!35}0.35&\cellcolor{blue!32}0.32&\cellcolor{blue!14}0.14&\cellcolor{blue!15}0.15&\cellcolor{blue!50}0.50&\cellcolor{blue!43}0.43&\cellcolor{blue!77}0.77&\cellcolor{blue!78}0.78&\cellcolor{blue!81}0.81&\cellcolor{blue!81}0.81\\
&&$\lambda=1$ &\cellcolor{Salmon!10}0.61 &\cellcolor{Salmon!0}0.46 &\cellcolor{Salmon!16}0.68 &\cellcolor{Salmon!5}0.55&\cellcolor{Salmon!32}0.90&\cellcolor{Salmon!30}0.87&\cellcolor{Salmon!18}0.69&\cellcolor{Salmon!18}0.70&\cellcolor{Salmon!42}1.04&\cellcolor{Salmon!37}0.97&\cellcolor{Salmon!60}1.31&\cellcolor{Salmon!60}1.32&\cellcolor{Salmon!60}1.34&\cellcolor{Salmon!60}1.33\\
&&$\lambda=10$ &\cellcolor{SeaGreen!15}2.05 &\cellcolor{SeaGreen!0}0.55 &\cellcolor{SeaGreen!18.6}2.48 &\cellcolor{SeaGreen!6}0.82&\cellcolor{SeaGreen!30}4.05&\cellcolor{SeaGreen!27}3.75&\cellcolor{SeaGreen!15}1.95&\cellcolor{SeaGreen!15}2.05&\cellcolor{SeaGreen!41}5.54&\cellcolor{SeaGreen!36}4.84&\cellcolor{SeaGreen!60}8.24&\cellcolor{SeaGreen!60}8.34&\cellcolor{SeaGreen!60}8.63&\cellcolor{SeaGreen!60}8.62\\
\cmidrule{2-17}
&\multirow{2}{*}{ACC ($\%$)} &$0.1$-T &\cellcolor{magenta!44}60.74 &\cellcolor{magenta!51}63.78 &\cellcolor{magenta!60}69.23 &\cellcolor{magenta!60}69.77&\cellcolor{magenta!67}73.36&\cellcolor{magenta!68}74.89&\cellcolor{magenta!73}76.86&\cellcolor{magenta!74}77.11&\cellcolor{magenta!77}79.50&\cellcolor{magenta!73}76.89&\cellcolor{magenta!82}81.15&\cellcolor{magenta!81}80.81&\cellcolor{magenta!81}80.71&\cellcolor{magenta!80}79.21\\
&&$0.01$-T &\cellcolor{magenta!00}34.03 &\cellcolor{magenta!06}37.71 &\cellcolor{magenta!10}40.16 &\cellcolor{magenta!17}44.67&\cellcolor{magenta!32}53.06&\cellcolor{magenta!41}58.11&\cellcolor{magenta!42}59.35&\cellcolor{magenta!50}63.08&\cellcolor{magenta!56}67.24&\cellcolor{magenta!58}68.50&\cellcolor{magenta!63}71.72&\cellcolor{magenta!66}72.85&\cellcolor{magenta!68}74.33&\cellcolor{magenta!67}73.54\\
\bottomrule
\multirow{6}{*}{\rotatebox{90}{DNN-Source}} &\multirow{4}{*}{SIM} &$\mathcal{V}$ &\cellcolor{cyan!60}0.60 &\cellcolor{cyan!60}0.60 &\cellcolor{cyan!43}0.55 &\cellcolor{cyan!30}0.53&\cellcolor{cyan!30}0.50&\cellcolor{cyan!27}0.50&\cellcolor{cyan!30}0.50&\cellcolor{cyan!21}0.49&\cellcolor{cyan!20}0.48&\cellcolor{cyan!20}0.48&\cellcolor{cyan!20}0.48&\cellcolor{cyan!17}0.47&\cellcolor{cyan!13}0.46&\cellcolor{cyan!10}0.45\\
&&$\mathcal{E}$ &\cellcolor{blue!06}0.06 &\cellcolor{blue!11}0.11 &\cellcolor{blue!15}0.15 &\cellcolor{blue!17}0.17&\cellcolor{blue!18}0.18&\cellcolor{blue!18}0.18&\cellcolor{blue!19}0.19&\cellcolor{blue!19}0.19&\cellcolor{blue!20}0.20&\cellcolor{blue!17}0.17&\cellcolor{blue!15}0.15&\cellcolor{blue!11}0.11&\cellcolor{blue!10}0.10&\cellcolor{blue!09}0.09\\
&&$\lambda=1$ &\cellcolor{Salmon!30}0.66 &\cellcolor{Salmon!60}0.71 &\cellcolor{Salmon!60}0.70 &\cellcolor{Salmon!60}0.70&\cellcolor{Salmon!40}0.68&\cellcolor{Salmon!40}0.68&\cellcolor{Salmon!50}0.69&\cellcolor{Salmon!35}0.67&\cellcolor{Salmon!40}0.68&\cellcolor{Salmon!20}0.65&\cellcolor{Salmon!20}0.63&\cellcolor{Salmon!15}0.58&\cellcolor{Salmon!15}0.56&\cellcolor{Salmon!10}0.54\\
&&$\lambda=10$ &\cellcolor{SeaGreen!0}1.20 &\cellcolor{SeaGreen!23}1.70 &\cellcolor{SeaGreen!37}2.05 &\cellcolor{SeaGreen!48}2.23&\cellcolor{SeaGreen!51}2.30&\cellcolor{SeaGreen!51}2.30&\cellcolor{SeaGreen!60}2.40&\cellcolor{SeaGreen!60}2.39&\cellcolor{SeaGreen!60}2.48&\cellcolor{SeaGreen!40}2.18&\cellcolor{SeaGreen!36}1.98&\cellcolor{SeaGreen!17}1.57&\cellcolor{SeaGreen!13}1.46&\cellcolor{SeaGreen!5}1.35\\
\cmidrule{2-17}
&\multirow{2}{*}{ACC ($\%$)} &$0.1$-T &\cellcolor{magenta!22}49.84 &\cellcolor{magenta!43}61.92 &\cellcolor{magenta!44}62.72 &\cellcolor{magenta!43}62.28&\cellcolor{magenta!42}59.81&\cellcolor{magenta!42.5}60.24&\cellcolor{magenta!37}58.49&\cellcolor{magenta!30}54.03&\cellcolor{magenta!30}54.21&\cellcolor{magenta!23}52.67&\cellcolor{magenta!23}52.15&\cellcolor{magenta!20}48.54&\cellcolor{magenta!8.3}41.50&\cellcolor{magenta!00}36.10\\
&&$0.01$-T &\cellcolor{magenta!00}30.58 &\cellcolor{magenta!20}35.49 &\cellcolor{magenta!28}37.20 &\cellcolor{magenta!36}39.47&\cellcolor{magenta!36}39.64&\cellcolor{magenta!36}39.63&\cellcolor{magenta!41}40.07&\cellcolor{magenta!41}40.11&\cellcolor{magenta!41}40.37&\cellcolor{magenta!36}39.04&\cellcolor{magenta!24}36.88&\cellcolor{magenta!16}34.13&\cellcolor{magenta!4}31.40&\cellcolor{magenta!0}29.13\\
\bottomrule
\end{tabular}
\end{center}
\end{table*}
\begin{figure}[t]
  \centering
  \subfigure[DNN-ImageNet.]{
  \begin{minipage}[t]{0.50\linewidth}
  \centering
  \includegraphics[scale=0.11]{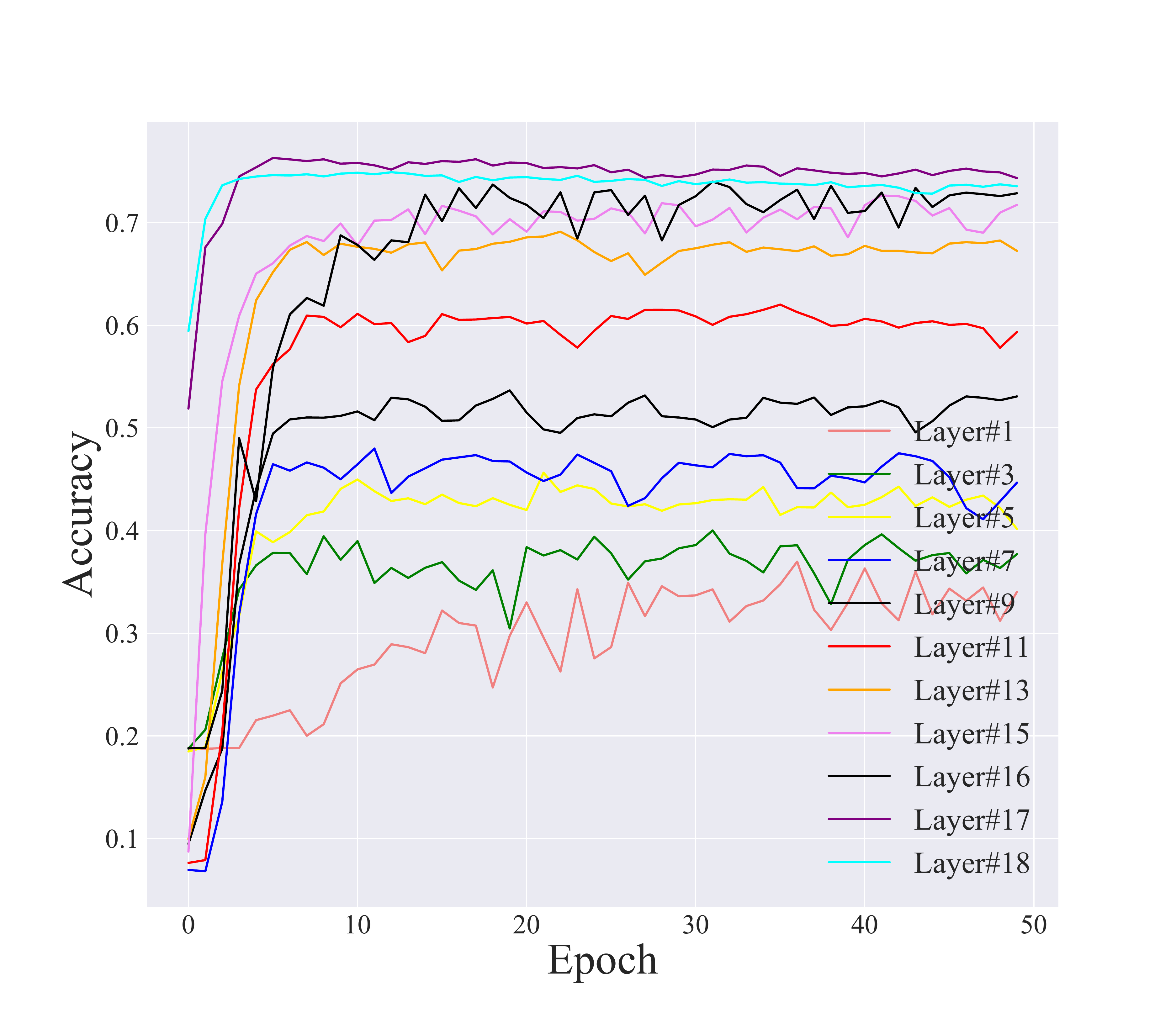}
  \end{minipage}%
  }%
  \subfigure[DNN-Source.]{
  \begin{minipage}[t]{0.50\linewidth}
  \centering
  \includegraphics[scale=0.11]{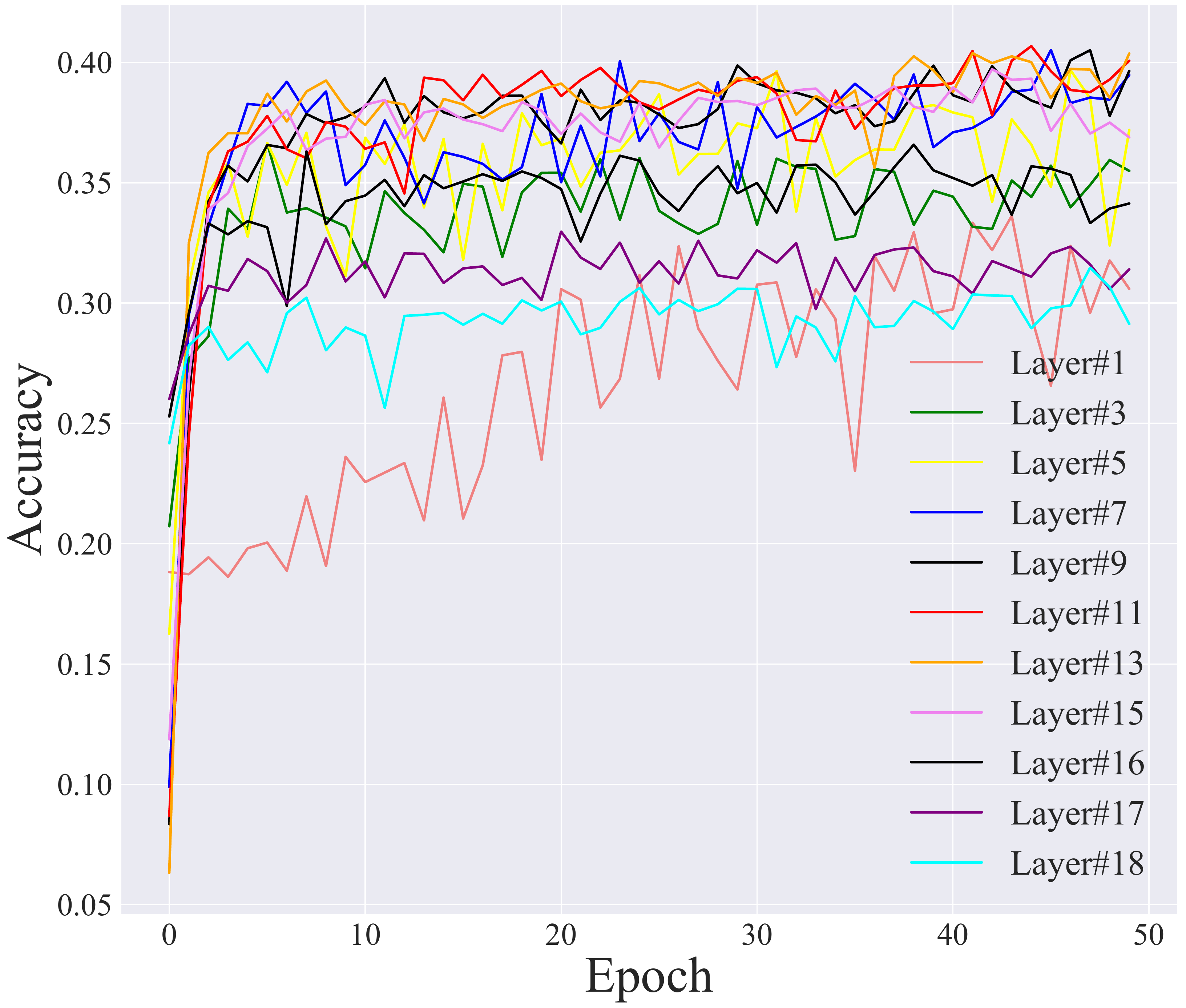}
  \end{minipage}%
  }%
  \caption{The test accuracy curves of different layers during the fine-tuning period in $0.01$-T mode.}
  \label{fig:accuracy_curve}
\end{figure}
Some other interesting observations from Table~\ref{table:layer_wise} are provided in the supplementary material. 

In Figure~\ref{fig:accuracy_curve}, we depict the test accuracy curves of different layers when transferred to the target data. The results further demonstrate the layers selected by the proposed method are more suitable for being transferred to the target than other layers. From Figure~\ref{fig:accuracy_curve}, it can be seen that the selected layers converge much faster than other layers when re-trained for the target task. For example, for the PR-DNN DNN-ImageNet, the proposed method picks out the $\#15$, $\#16$, $\#17$, $\#18$ layers for being transferred. The final accuracy also tends to be higher than that of other layers. Furthermore, layers in DNN-ImageNet produce more smooth test accuracy curves than DNN-Source, which indicates that the embedding space learned by DNN-ImageNet are more easily adapted to the target task. The embedding space learned by DNN-Source, however, is quite different in topological structure (as indicated by the low similarity of edges in DEPARA) from the one learned on the target data. When adapted to the target data, it will be largely destroyed and rebuilt for the target, thus the test accuracy curves oscillate and the transferring performance is poor.

\section{Discussions and Conclusions}
In this paper, we propose the DEPARA to investigate the transferability of knowledge encoded in PR-DNNs. We adopt DEPARA to handle two important yet under-studied problems in transfer learning: measuring the transferability across tasks for pre-trained model selection, and measuring the transferability across layers for layer selection. Extensive experiments are conducted to show its effectiveness in solving both these two problems in transfer learning. We summarize the advantages and the limitations of the proposed method. We hope it could make the contributions of this paper clearer and inspire us to study further.

\textbf{Advantages.} (1) Unlike taskonomy~\cite{Zamir_2018_CVPR} which requires a large amount of labeled data, the proposed method quantifies the task transferability with only pre-trained models available. (2) As no training is involved, the computation cost of the proposed method grows nearly linearly with the size of the task dictionary, which is significantly more efficient than taskonomy. (3) The proposed method solves not only the model selection, but also the layer selection problem. As far as we know, we are the first to simultaneously tackle the model and the layer selection problems in transfer learning. (4) The proposed method imposes no constraints on the model architectures and are insensitive to the probe data. (5) This paper introduces a rigorous definition of knowledge transferability. Meanwhile, two vital ingredients, including inclusiveness and accessibility, are introduced for better approximating the transferability.

\textbf{Limitations.} (1) This paper directly adopts the existing attribution method, Gradient*Input~\cite{Shrikumar2016NotJA}, for quantifying transferability. However, different attribution methods may affect the proposed method in some way. In future work, more studies are needed to investigate the effects of different attribution methods on the proposed method. (2) The optimal trade-off between the nodes and the edges of DEPARA for knowledge transferability is proved to be dependent on the probe data and the amount of the target data. In this paper, the trade-off hyper-parameter $\lambda$ is set via cross-validation or empirically. However, more study is needed to uncover the relationship between $\lambda$ and its influencing factors. (3) The probe data used in the proposed method is randomly sampled. Although different probe data are shown to produces effective task-wise topological structures, they still affect the final performance to some degree. More investigation is needed to study how to construct the probe data for better measuring the transferability across different tasks, models and layers.

\paragraph{Acknowledgments.} This work is supported by  National Key Research and Development Program (2018AAA0101503), National Natural Science Foundation of China (61976186), Key Research and Development Program of Zhejiang Province (2018C01004), and the Major Scientific Research Project of Zhejiang Lab (No. 2019KD0AC01).

{\small
\bibliographystyle{ieee_fullname}
\bibliography{cvpr_2020}
}

\end{document}